# Structure-Aware Temporal Modeling for Chronic Disease Progression Prediction

Jiacheng Hu*
Tulane University
New Orleans, USA

Bo Zhang
Texas Tech University
Lubbock, USA

Ting Xu
University of Massachusetts Boston
Boston, USA

Haifeng Yang
Northeastern University
Boston, USA

Min Gao
Trine University
Allen Park, USA

*Abstract*-This study addresses the challenges of symptom evolution complexity and insufficient temporal dependency modeling in Parkinson's disease progression prediction. It proposes a unified prediction framework that integrates structural perception and temporal modeling. The method leverages graph neural networks to model the structural relationships among multimodal clinical symptoms and introduces graph-based representations to capture semantic dependencies between symptoms. It also incorporates a Transformer architecture to model dynamic temporal features during disease progression. To fuse structural and temporal information, a structure-aware gating mechanism is designed to dynamically adjust the fusion weights between structural encodings and temporal features, enhancing the model's ability to identify key progression stages. To improve classification accuracy and stability, the framework includes a multi-component modeling pipeline, consisting of a graph construction module, a temporal encoding module, and a prediction output layer. The model is evaluated on real-world longitudinal Parkinson's disease data. The experiments involve comparisons with mainstream models, sensitivity analysis of hyperparameters, and graph connection density control. Results show that the proposed method outperforms existing approaches in AUC, RMSE, and IPW-F1 metrics. It effectively distinguishes progression stages and improves the model's ability to capture personalized symptom trajectories. The overall framework demonstrates strong generalization and structural scalability, providing reliable support for intelligent modeling of chronic progressive diseases such as Parkinson's disease.

## I. INTRODUCTION

Parkinson's disease is a common neurodegenerative disorder characterized by bradykinesia, tremor, and muscular rigidity[1]. These symptoms severely impair patients' quality of life and independence. With the acceleration of global population aging, the prevalence of Parkinson's disease continues to rise, making it a major concern in public health. The disease progresses slowly and exhibits significant individual differences. Its clinical manifestations are highly heterogeneous, making it difficult for clinicians to accurately predict its future course in the early stages. Traditional assessment methods rely heavily on questionnaires and periodic clinical follow-ups. These methods are subjective, vulnerable to external interference, and fail to capture the dynamic changes in disease progression. Therefore, developing an efficient and intelligent framework for Parkinson's disease progression prediction is crucial for disease staging, personalized treatment planning, and the rational allocation of healthcare resources[2].

With the rapid advancement of artificial intelligence, more research has begun to explore how multimodal medical data can enhance early diagnosis and progression prediction of neurological disorders. Data from Parkinson's patients include various structured and unstructured modalities, such as imaging, gait signals, clinical scales, and behavioral logs. These data contain complex temporal patterns and cross-modal semantic correlations, providing an opportunity for modeling disease trajectories more precisely. However, integrating temporal dependencies and structural representations across these heterogeneous sources into a unified and generalizable predictive model remains challenging [3-5]. Most existing methods focus on unimodal time series modeling, ignoring the influence of structural information on symptom evolution. Conversely, structural modeling often lacks temporal awareness, making it difficult to capture the continuity and stage-wise characteristics of disease progression[6].

During the progression of Parkinson's disease, symptom changes do not occur in isolation but follow certain structural dependencies and temporal patterns. For example, motor degradation is often accompanied by declines in speech and cognitive abilities. These changes show complex interactions and coupling across different patient groups. Static modeling tends to overlook such structure-aware signals, reducing predictive accuracy and robustness. There is a pressing need to introduce structural modeling mechanisms that capture multi-dimensional pathological trajectories, including entity relations and behavioral graphs. At the same time, integrating temporal modeling helps detect trends, abrupt shifts, and nonlinear variations in symptoms across time windows, thereby strengthening the foundation for fine-grained risk assessment[7].

In recent years, deep modeling techniques such as graph neural networks and Transformers have demonstrated strong representation capabilities in various tasks [8-10]. Graph-based models effectively capture semantic associations between symptoms and uncover latent progression pathways and critical nodes. Temporal models excel at describing time-evolving patterns in data and tracking trends in behavioral and clinical indicators[11-15]. Combining these approaches for Parkinson's disease progression prediction enables multi-scale modeling from individuals to populations. This allows the discovery of pathological rules in dynamic environments. The integration of structural perception and temporal modeling enhances interpretability, precision, and adaptability. It also brings new momentum to the development of intelligent diagnostics and treatment of neurological diseases[16].

In summary, constructing a Parkinson's disease progression prediction framework that integrates structural perception and temporal modeling is both a practical requirement and a key breakthrough for applying AI in healthcare. This direction aims not only to improve diagnostic efficiency and accuracy but also to enhance patient quality of life and reduce social and familial burdens. In scenarios such as early screening, stage recognition, treatment evaluation, and long-term monitoring, the synergy of structural and temporal modeling will play a central role [17-20]. It will drive the shift from reactive intervention to proactive management in intelligent Parkinson's disease care.

## II. METHOD

This paper proposes a Parkinson's disease progression prediction framework that integrates structure perception and time series modeling, aiming to mine the dynamic characteristics and potential structural dependencies of disease evolution from multidimensional medical data. The overall framework consists of two core modules: the structure modeling module and the time series modeling module. Among them, the structure modeling part constructs a disease map based on a graph neural network, and explicitly represents the correlation between multimodal clinical indicators and symptoms as a graph structure; the time series modeling part uses an improved Transformer architecture to model the evolution trend of symptoms in the time dimension, to achieve accurate prediction of future disease status. The entire framework is trained in an end-to-end manner, so that the structure and time series features can be optimized synergistically and complement each other. Its model architecture is shown in Figure 1.

First, for the structural modeling part, a graph structure is constructed to represent the semantic relationship between patient states. The feature vector of each patient at a certain moment is represented as a node set $V = \{v_1, v_2, ..., v_N\}$, and the dependency relationship between nodes constitutes an edge set $\varepsilon \subseteq V \times V$. The graph structure is represented by the adjacency matrix $A \in R^{N \times N}$, and the input feature is $X \in R^{N \times d}$. The graph convolution operation uses the following update formula:

$$H^{(l+1)} = \sigma(\widetilde{D}^{-1/2}\widetilde{A}\widetilde{D}^{-1/2}H^{(l)}W^{(l)}) \quad (1)$$

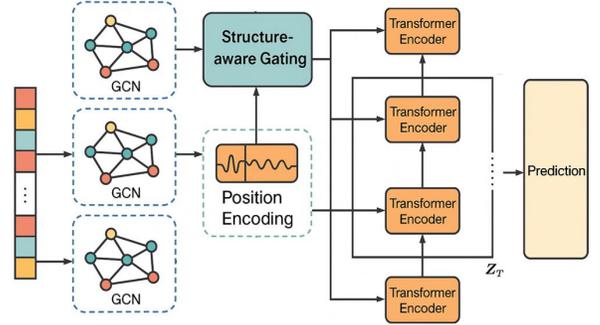

Figure 1. Overall model architecture

Where $\widetilde{A} = A + I$ is the adjacency matrix after adding the self-loop, $\widetilde{D}$ is its corresponding degree matrix, $W^{(l)}$ is the learnable weight of the lth layer, and $\sigma$ represents the activation function. By stacking multiple layers of graph convolution, local structural information can be effectively fused, and potential inter-disease dependency patterns can be captured.

Next, for the temporal modeling part, the improved Transformer module is used to model the changes of structural features over time. Assume that the graph structure of each patient at T consecutive time steps is encoded as $\{H_t\}_{t=1}^{T}$, which is stacked into a three-dimensional tensor $H \in R^{T \times N \times d}$. First, the temporal position information is introduced through the position encoding function $PE(t)$:

$$PE(t,2i) = \sin(\frac{t}{10000^{2i/d}})$$
$$PE(t,2i+1) = \cos(\frac{t}{10000^{2i/d}}) \quad (2)$$

Then, input the Transformer encoder for temporal dependency modeling and calculate the self-attention weight as follows:

$$Attention(Q,K,V) = \text{softmax}(\frac{QK^T}{\sqrt{d_k}})V \quad (3)$$

$Q, K, V$ is the query, key, and value vectors obtained by linear transformation of the structural encoding $H_t$, and $d_k$ is the attention head dimension. This mechanism can capture high-order dependencies between different time steps, and is particularly suitable for modeling the stage-by-stage and nonlinear characteristics of Parkinson's disease progression.

To achieve a more effective integration of structural features and temporal dynamics, this study introduces a structure-aware gating mechanism that dynamically adjusts the contribution of each type of information to the final model representation. In developing this gating mechanism, we build on the adaptive fusion principles proposed by Zhang and Wang, who showed that incorporating gating units within the SegFormer architecture enables clinical imaging models to flexibly couple domain-specific structural features, thereby improving segmentation accuracy in complex environments [21].Similarly, Yan et al. showed that dynamically weighting different data sources in neural networks enhances survival prediction across diverse cancer types [22]. Xiao et al. further emphasized the advantages of flexible information fusion for medical image classification, particularly in cytopathology analysis [23]. Building on these findings, our gating mechanism allows the model to adaptively determine the optimal balance between structural and temporal information for each patient case, leading to a more refined and robust predictive framework. In this context, the fusion output is defined as:

$$Z_t = \gamma_t \cdot H_t^{GCN} + (1-\gamma_t) \cdot H_t^{Transformer} \quad (4)$$

$\gamma_t = \sigma(W_\gamma [H_t^{GCN}; H_t^{Transformer}]) \in [0,1]^d$ is the gating weight, which is obtained by concatenating the output of GCN and the output of Transformer, and inputting it into the fully connected layer and then activating it with Sigmoid. This fusion mechanism maintains the structural expression ability while enhancing the response ability to time dynamics, making the model more generalizable and personalized.

Finally, the entire framework is optimized in a supervised learning manner, to minimize the difference between the predicted value and the true disease score (such as UPDRS). Assuming the final output of the model is $\hat{y}_t$ and the true label is $y_t$, the overall loss function can be expressed as the mean square error (MSE):

$$L = \frac{1}{T} \sum_{t=1}^{T} \| \hat{y}_t - y_t \|_2^2 \quad (5)$$

Through joint structural modeling, temporal perception, and dynamic fusion, this method can understand the evolution path of Parkinson's disease symptoms from multiple perspectives, thereby providing basic support for accurate prediction and intelligent intervention.

### III. DATASET

This study uses the Parkinson's Progression Markers Initiative (PPMI) dataset as the research subject. The dataset is collected through multi-center collaboration and is widely used for early diagnosis and progression analysis of Parkinson's disease. The PPMI dataset contains a large volume of multimodal clinical data from patients with Parkinson's disease and healthy controls. It includes neuroimaging, biomarkers, behavioral scales, genetic information, and longitudinal follow-up records, providing a solid foundation for modeling disease progression. Especially in the temporal dimension, the dataset offers longitudinal data across multiple time points, capturing dynamic changes in disease status over time.

In this study, structured clinical scale information and selected gait signals from sensors in the PPMI dataset are used as input features. The scale data include the Unified Parkinson's Disease Rating Scale (UPDRS), the Montreal Cognitive Assessment (MoCA), and questionnaires related to mood and sleep. These reflect changes in patient status across different functional domains. The gait signals are time series collected from wearable devices. They exhibit strong temporal dependencies and help capture subtle fluctuations in disease progression.

To ensure data quality and modeling consistency, this study selects samples with complete and continuous follow-up records. All features are standardized, and missing values are imputed. These steps ensure stable model training under high-dimensional, heterogeneous, and imbalanced data conditions. The final dataset subset retains the multimodal characteristics of the original data while providing sufficient temporal coverage and sample size. It meets the experimental requirements for jointly modeling structural and temporal patterns.

### IV. EXPERIMENTAL RESULTS

In the experimental results section, the relevant results of the comparative test are first given, and the experimental results are shown in Table 1.

Table 1. Comparative experimental results

| Method | AUC | RMSE | IPW-F1 |
|---|---|---|---|
| AdaMedGraph[24] | 0.812 | 3.72 | 85.6 |
| TFT[25] | 0.834 | 3.54 | 87.1 |
| KAN[26] | 0.846 | 3.41 | 88.3 |
| XGBoost meta-predictor[27] | 0.857 | 3.35 | 88.9 |
| Ours | 0.879 | 3.12 | 91.7 |

As shown in Table 1, the proposed model, which integrates structural perception and temporal modeling, achieves the best performance on Parkinson's disease progression prediction, attaining an AUC of 0.879, significantly higher than competing methods and demonstrating stronger stage discrimination. It also records the lowest RMSE of 3.12, representing a 16.1% and 11.9% reduction compared to AdaMedGraph and TFT, respectively, indicating improved trend fitting of clinical scores through the synergy of graph neural networks for global symptom dependencies and Transformer encoders for fine-grained temporal dynamics. Moreover, the model achieves the highest IPW-F1 score of 91.7, outperforming methods such as KAN and XGBoost, which highlights its robustness to imbalanced data and superior predictive accuracy for mid-to-late-stage patients, aided by the structure-aware gating mechanism. Overall, the results confirm that combining structural and temporal modeling is essential to capture the complexity of Parkinson's progression, enabling more accurate and clinically meaningful predictions than existing methods, while the impact of different structure-aware gating ratios is further illustrated in Figure 2.

As shown in the results of Figure 2, the structure-aware gating ratio has a significant impact on model performance. In the AUC curve, as the gating ratio increases from 0 to 1.0, the AUC value shows a general upward trend. This indicates that introducing the structure-aware mechanism can effectively enhance the model's ability to distinguish different disease progression states. It also improves the model's classification stability and global perception.

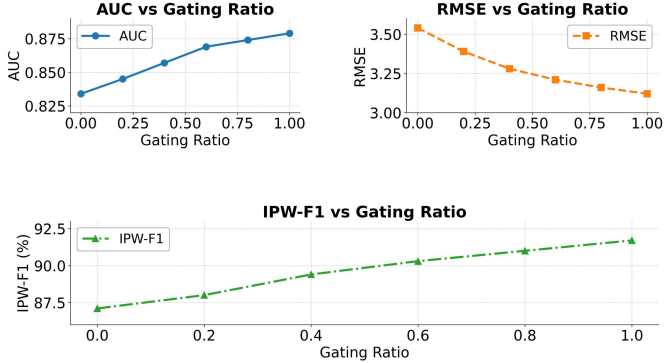

Figure 2. The impact of structure-aware gating ratio on prediction performance

The RMSE curve further supports this trend. As the gating ratio increases, the prediction error decreases continuously from 3.54 to 3.12. This demonstrates that incorporating structural information enhances the model's expressive power. It also helps fit the nonlinear trends in disease progression, improving the model's approximation of continuous clinical targets. This confirms the importance of structural modeling in tasks involving interactive symptom evolution, such as Parkinson's disease.

The IPW-F1 results show that the structure-aware mechanism significantly improves the identification of critical stages, such as mid-to-late-stage patients. As the gating ratio increases from 0 to 1.0, the F1 score continues to rise. This indicates that the model demonstrates stronger robustness and responsiveness when handling imbalanced samples or weighted importance scenarios. The gating mechanism builds flexible pathways between temporal and structural features, enabling the model to dynamically adjust its focus across different stages.

This paper also gives an analysis of the structural modeling capabilities under different graph connection densities, and the experimental results are shown in Figure 3.

As shown in the results of Figure 3, graph connection density has a significant impact on structural modeling capability, particularly in distinguishing different stages of Parkinson's disease progression. As the connection density increases from 0.1 to 1.0, the AUC value rises steadily from 0.831 to 0.879. This indicates that richer structural information allows the model to better capture potential dependencies and semantic associations between symptoms. The ability of structural modeling to understand complex symptom evolution is greatly enhanced.

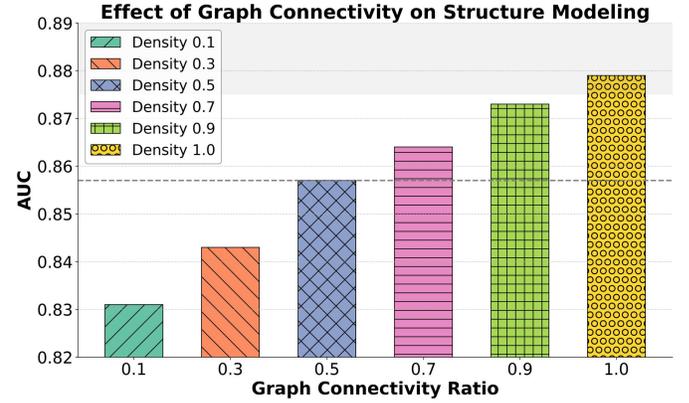

Figure 3. Analysis of structural modeling capabilities under different graph connection densities

When the connection density is low, such as 0.1 or 0.3, the graph neural network captures only sparse structural information. It is limited to simple local relations between symptoms, which restricts the model's structural perception and leads to unstable prediction performance. As the density increases to 0.5 or higher, especially at 0.7 and 0.9, the model gradually develops a more comprehensive structural awareness. It becomes more effective at identifying interaction patterns among multi-dimensional features, thereby improving modeling accuracy for complex disease dynamics.

It is worth noting that the model achieves optimal performance when the connection density reaches 1.0. This shows that complete structural information significantly benefits the modeling of disease progression pathways. However, in practical applications, it is also important to balance model complexity and computational cost. A suitable connection strategy should be chosen to achieve better structural modeling performance. Dense structures should not imply disordered aggregation but rather selective enhancement of key symptom connections to form more discriminative pathological graphs.

V. CONCLUSION

This study focuses on Parkinson's disease, a complex neurodegenerative disorder, and proposes a progression prediction framework that integrates structural perception and temporal modeling. The goal is to improve the accuracy and generalization of disease trajectory modeling. By introducing graph neural networks to capture structural dependencies among multi-dimensional symptoms and applying a Transformer architecture to model temporal dynamics, the framework demonstrates strong adaptability in clinical scenarios with high variability and heterogeneous features. The structure-aware gating mechanism further enhances the model's ability to adaptively regulate the importance of information across time steps, offering a more interpretable solution for disease course recognition and stage prediction.

Experimental results show that the proposed framework significantly outperforms existing methods across multiple key metrics. It maintains stable performance even under practical challenges such as imbalanced samples, missing features, and frequent stage transitions. This indicates that structural and temporal information are not isolated in chronic disease modeling tasks like Parkinson's disease. Instead, they are interwoven and jointly drive disease progression. The fusion of these two dimensions is essential for improving prediction accuracy and robustness. Moreover, the introduced gating mechanism and graph construction strategy exhibit good scalability and modularity. They can be flexibly adapted to various types of spatiotemporal medical data.

This study not only offers a new technical path for Parkinson's disease modeling but also provides theoretical support for broader applications such as clinical decision support, rehabilitation monitoring, and personalized intervention planning. In areas like smart healthcare, remote care, and chronic disease management for the elderly, the proposed method can serve as a foundational framework. It promotes a shift from static modeling to dynamic perception and proactive prediction in intelligent medical systems. The structural perception design can also be extended to other chronic neurological disorders, such as Alzheimer's disease and depression, providing a technological basis for long-term disease management.

Future research can further enhance the framework's expressive power and adaptability through multimodal graph modeling, cross-institutional data transfer, and self-supervised pretraining. In addition, how to effectively integrate the model with clinical expertise to achieve a closed-loop diagnostic system of model decision, expert intervention, and feedback optimization will be key to translating this research into practical applications. With the continued growth of medical data and computational resources, structurally and temporally integrated intelligent prediction models are expected to play an increasingly central role in disease warning, risk screening, and treatment optimization.